\documentclass[conference]{IEEEtran}
\IEEEoverridecommandlockouts
\IEEEpubid{\makebox[\columnwidth]{DOI: 10.1109/IJCNN52387.2021.9533762 ©2021 IEEE \hfill} \hspace{\columnsep}\makebox[\columnwidth]{ }}
\usepackage{cite}
\usepackage{amsmath,amssymb,amsfonts}
\usepackage[ruled,vlined]{algorithm2e}
\usepackage{array}
\usepackage{multirow}
\usepackage{url}
\usepackage[export]{adjustbox}
\usepackage{graphicx}
\usepackage{textcomp}
\usepackage[dvipsnames]{xcolor}
\def\BibTeX{{\rm B\kern-.05em{\sc i\kern-.025em b}\kern-.08em
    T\kern-.1667em\lower.7ex\hbox{E}\kern-.125emX}}
\begin{document}

\title{Learning Sparsity of Representations with \\
Discrete Latent Variables
\thanks{The work of NEC Laboratories Europe was partially supported by H2020 MonB5G project (grant agreement no. 871780). The research of G. Serra was supported by H2020 ECOLE project (grant agreement no. 766186).}
}

\author{\IEEEauthorblockN{Zhao Xu\textsuperscript{\dag}, Daniel Onoro Rubio\textsuperscript{\dag}, Giuseppe Serra\textsuperscript{\dag\S}, Mathias Niepert\textsuperscript{\dag}}

\IEEEauthorblockA{\textsuperscript{\dag}{NEC Laboratories Europe, Heidelberg, Germany} \\
\textsuperscript{\S}{CERCIA, School of Computer Science, University of Birmingham, UK}\\
\{zhao.xu, daniel.onoro, giuseppe.serra, mathias.niepert\}@neclab.eu}
}

\maketitle

\begin{abstract}
Deep latent generative models have attracted increasing attention due to the capacity of combining the strengths of deep learning and probabilistic models in an elegant way. The data representations learned with the models are often continuous and dense. However in many applications, sparse representations are expected, such as learning sparse high dimensional embedding of data in an unsupervised setting, and learning multi-labels from thousands of candidate tags in a supervised setting. In some scenarios, there could be further restriction on degree of sparsity: the number of non-zero features of a representation cannot be larger than a pre-defined threshold $L_0$. In this paper we propose a sparse deep latent generative model SDLGM to explicitly model degree of sparsity and thus enable to learn the sparse structure of the data with the quantified sparsity constraint. The resulting sparsity of a representation is not fixed, but fits to the observation itself under the pre-defined restriction. In particular, we introduce to each observation $i$ an auxiliary random variable $L_i$, which models the sparsity of its representation. The sparse representations are then generated with a two-step sampling process via two Gumbel-Softmax distributions. For inference and learning, we develop an amortized variational method based on MC gradient estimator. The resulting sparse representations are differentiable with backpropagation. The experimental evaluation on multiple datasets for unsupervised and supervised learning problems shows the benefits of the proposed method.
\end{abstract}

\section{Introduction}
\label{sec:intro}

Deep Latent Generative Models (DLGMs) \cite{Kingma2014,Rezende2014,Gregor2014,Sohn2015,Jang2017} are powerful generative models that combines strength of deep neural networks and approximate Bayesian inference to learn representations of large scale data. The learned representations reveal important semantic features of the data, and are often useful for many downstream tasks. The DLGMs have been successfully applied to a variety of applications with interesting results, such as question-answering systems \cite{Miao2016}, action forecasting \cite{Walker2016}, anomaly detection \cite{Xu2018}, image progression and editing \cite{Yan2016,Yeh2016}.

In this paper, we investigate sparse deep latent generative model (SDLGM) to explicitly learn sparsity of discrete representations of the data. The semantic features of large scale data are often sparse. For instance, in an image database with a variety of images, possible visual concepts include hairstyles, pose and emotions for portraits, art-styles for paintings, animals and plants for nature etc. Different types of images will only focus on a limited number of latent semantic features. Thus the latent codes of the images should be intrinsically sparse. Another typical example in a supervised learning setting might be social tags of recommendation systems. For example, the total number of candidate social tags of an online music recommendation system can be hundreds of thousands, but a single track is often labeled with a dozen of tags. Thus it makes sense to learn sparse discrete representations such that the learned logits mainly focus on a finite number of features. Moreover, there could be further constraint on sparsity in some scenarios. For example, the number of non-zero features of an observation (image or time series) cannot be larger than a pre-defined threshold $L_0$ for IoT devices with limited resources in terms of storage and communication bandwidth. 

To meet these challenges, the SDLGM introduces to each observation an auxiliary (discrete) random variable $L_i$, which specifies how sparse the representation should be. The smaller the variable $L_i$, the less number of features are required by the observation to represent its intrinsic properties. Then we assume the representation is a sum of $L_i$ i.i.d. features from the shared feature distributions. In a Bayesian framework, it would be natural to model $L_i$ as a discrete variable with e.g.~binomial distribution. However it makes the feature sampling process expensive, since the number of random features is varying. Technically, we introduce to each observation $L_0$ i.i.d. random features with the corresponding i.i.d. Bernoulli weights. The weights can be viewed as gates. When a weight is zero, then the gate switches off, and the corresponding feature is dropped out. The resulting representation is a weighted sum of $L_0$ random features. Moreover the sum of the weights is exactly $L_i$ following binomial distribution. Under the deep learning framework, the discrete sparsity is unfortunately non-differentiable, and in turn makes backpropagation impossible. To solve the problem, we employ a Gumbel softmax distribution \cite{Jang2017,Maddison2017,Tucker2017}, one for each observation and conditioned on the observation itself via a DNN, as the distinct sparsity distribution. This is a differentiable approximation to its discrete counterpart. For feature distributions, we employ high dimensional Gumbel-Softmax distribution. 
In the SDLGM, the sparsity of the representation is explicitly modeled with an extra random variable, and thus can be clearly learned from the data itself. It is distinct from the commonly used regularization techniques, which implicitly achieve sparsity effect. The proposed two-step sampling process allows the SDLGM method efficiently capture significant features of the observations. It nicely matches the context where there is a quantified restriction on the sparsity itself.

We apply the SDLGM to supervised and unsupervised learning cases. For supervised learning, the problem is formulated as a multi-class multi-label classification issue. We exploit three modeling strategies, including discriminative modeling (SDLGM-Dis), generative modeling (SDLGM-Gen) and conditional generative modeling (SDLGM-Con).  
In the unsupervised case, the problem is the representation learning, where the representation variable $\mathbf{z}_i$ is unobservable. We learn the joint probability $P(\mathbf{z}_i,\mathbf{x}_i)=P(\mathbf{z}_i)P(\mathbf{x}_i|\mathbf{z}_i)$.  
The inference of the SDLGM is based on amortized variational method and MC gradient estimator. 
The empirical analysis is performed on multiple datasets. We compare the proposed model with the most recent results. For unsupervised learning, we use the popular benchmark dataset MNIST, and measure the performance with a commonly used criteria: lower bound of the log likelihood of the data. The SDLGM significantly outperforms the baselines. In the qualitative analysis, the proposed model shows very clear reconstructed images and latent structures of the data. For supervised learning, the SDLGM method also achieves better performance in terms of Micro- and Macro-averaging F1 scores in multiple datasets: Birds, Emotions and CAL500. 

The rest of the paper is organized as follows. We start off with a brief review of related work. Afterwards we describe the proposed model SDLGM to learn the sparsity of representation and the variational inference method. Before concluding, we present our experimental results on multiple datasets for different learning tasks.

\section{Related Work}
\label{sec:relatedwork}

There are mainly two lines of research related to the proposed  model SDLGM, including Bayesian generative neural networks and sparse representation learning. In this section, we briefly review the related work.

\subsection{Bayesian Generative Neural Networks}
Recent advances in Bayesian modeling renew interests in scalable probabilistic modeling. 
There are two major benefits which make Bayesian methods appealing: modeling uncertainty in the learned parameters and avoiding overfitting. 
They compensate some pitfalls of deep neural networks. For example, with increasingly complex problems and large data, the deep neural networks become more and more complicated and often include millions of parameters. So it is required to quantify the amounts of parameter uncertainties.

Introducing Bayesian modeling into neural networks have attracted a lot of attention in the literature. The early works include: for example, Denker and LeCun transformed neural network outputs to probabilistic distributions \cite{Denker1991}. Buntine and Weigend presented Bayesian back-propagation \cite{Buntine1991}. Neal proposed the first Monte Carlo (MC) sampling method to learn Bayesian neural networks \cite{Neal1993}. 

With the emergence of deep learning, scalability of Bayesian methods become a major concern.
Recent Bayesian methods incorporate the advances in optimization, and enable Bayesian models for large scale datasets. For instance, Hoffman et al.~introduced stochastic variational inference method \cite{Hoffman2013}. Welling and Teh presented stochastic gradient Langevin dynamics to approach the true posterior distribution of the parameters \cite{Welling2011}. The Bayesian backpropagation was extended to large scale data \cite{Graves2011,Blundell2015}. 

Deep latent generative models DLGMs nicely integrate recent advances of Bayesian modeling into deep neural networks for large data. The DLGMs are originally developed to learn continuous latent representations for unsupervised learning. The pioneer works include variational autoencoder \cite{Kingma2014}, deep latent Gaussian models \cite{Rezende2014}, and autoregressive networks \cite{Gregor2014}. The models are then extended to semi-supervised learning in \cite{Kingma2014b}. Sohn et al.~presented conditional DLGMs to learn structured representations. The discrete version of the DLGMs are introduced by \cite{Jang2017,Maddison2017,Tucker2017} to learn discrete latent representations with Gumbel-Softmax. The DLGMs are also extended to model text data \cite{Miao2016}, graphical data \cite{Kipf2016} and multimodal data \cite{Pu2016}.

\subsection{Sparse Representation with Neural Networks}
Sparse representation with neural networks has been investigated from different perspectives in the literature. 

Lee et al.~\cite{Lee2008} and Ranzato et al.~\cite{Ranzato2008} introduced sparse variants of deep belief networks to learn sparse features of images.
Sparse auto-encoders were explored by Kavukcuoglu et al.~\cite{Kavukcuoglu2010}, Makhzani and Frey \cite{Makhzani2014}, and Arpit et al.~\cite{Arpit2016} for sparsification of neural network layers.

Glorot et al.~\cite{Glorot2011} were motivated by certain properties of biological neurons and proposed rectifier activation function to  sparsify neural networks.
Goodfellow et al.~\cite{Goodfellow2013}, Martins and Astudillo \cite{Martins2016} introduced Maxout and SparseMax activation functions for sparsity, respectively.
Dropout method has also been used to effectively sparsify neural networks by Molchanov et al.~\cite{Molchanov2017}.
Louizos et al.~\cite{Louizos2018} and Srinivas et al.~\cite{Srinivas2016} investigated $\ell_0$ norm related regularization methods for learning sparse neural networks.

Deep sparse representation is also developed for fast similarity search. For example, Li et al. introduced the optimal sparse lifting method \cite{Li2018}.  Jeong and Song presented an optimization method to jointly learn a quantizable embedding representation and the sparse hash table\cite{Jeong2018}.
Inspired by efficient sparse matrix multiplication, Paria et al. introduced to learn sparse representation by minimizing a continuous relaxation of the number of floating-point operations (FLOPs) \cite{Paria2020} in search.

Distinguishing from the existing work, this paper extends the deep latent generative models to explicitly quantity  degree of  sparsity, and fit it to the data automatically. We also apply the learnable sparsity to supervised learning for multi-label classification with good performance.

\section{Sparse Deep Latent Models}
\label{sec:model}

\begin{figure}
\centering
\includegraphics[width=7.5cm]{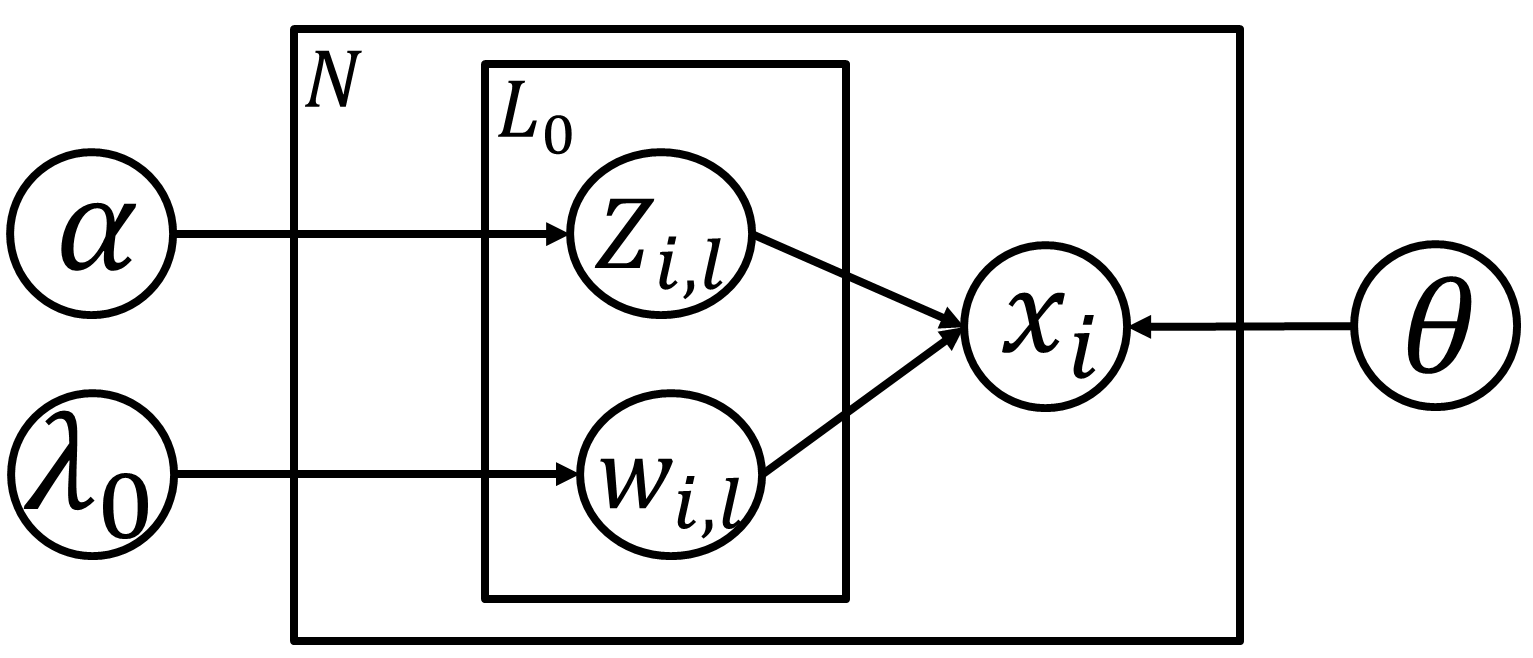}
\caption{Sampling process of the sparse deep latent generative model SDLGM.}
\label{fig:model}
\vspace{-0.3cm}
\end{figure}
In this section, we introduce the proposed sparse deep latent generative model SDLGM and the corresponding inference method. We will first describe the model in an unsupervised learning case to learn representations of the data, then extend it to a supervised learning case for multi-class multi-label classification. 
Let start with problem formulation. Assume that there are $N$ observations $\mathbf{X} = \{\mathbf{x}_1,\mathbf{x}_2,\ldots,\mathbf{x}_N\}$. Each observation $\mathbf{x}_i$ is a $D$-dimensional vector. We assume that 
$x_{i,d}$ is binary for the running example, while it is straightforward to extend the SDLGM to continuous inputs. The goal is to learn a sparse $K$-dimensional discrete representation $\mathbf{z}_i$ for each observation under a constraint on representation sparsity that the number of non-zero features cannot be larger than a certain threshold $L_0$.
Formally the representation $\mathbf{z}_i$ of the observation $\mathbf{x}_i$ is denoted as:
$
\mathbf{z}_i \in \{0,1\}^K; \;\; ||\mathbf{z}_i||_0\leq L_0
$,
where $||\cdot||_0$ is  $\ell_0$ norm of a vector.

The SDLGM is a probabilistic generative model which integrates good properties of deep neural networks DNNs into Bayesian modeling for  flexible definition of data distributions. 
To meet the sparsity constraint, we introduces to each observation $\mathbf{x}_i$ an auxiliary discrete random variable $L_i$, which reveals degree of representation sparsity.  The value of the variable $L_i$ specifies how many non-zero features are required by the observation to represent its intrinsic properties. In addition, $L_i$ corresponds to the $\ell_0$ norm of $\mathbf{z}_i$. Technically it is tricky to directly model the $\ell_0$ norm $L_i$, since the varying number of random features makes the sampling process inefficient. Alternatively, we decompose the representation $\mathbf{z}_i$ as:
$
\mathbf{z}_i = \sum\nolimits_{\ell=1}^{L_0} \omega_{i,\ell} \mathbf{z}_{i,\ell}
$.
It specifies that the representation $\mathbf{z}_i$ consists of at most $L_0$ features. Each feature $\mathbf{z}_{i,\ell}$ is a $K$-dimensional one-hot vector, and corresponds to a categorical variable. The weight $\omega_{i,\ell}$ is binary, and can be viewed as a \textit{gate}. When the gate switches on $\omega_{i,\ell} =1$, the corresponding feature is present. The gate vector $\omega_{i}=(\omega_{i,1},\ldots,\omega_{i,L_0})$ is observation-specific, one for each observation. It will be learned as per the data itself. The sparsity variable can thus be written as: 
$
L_i=\sum\nolimits_{\ell=1}^{L_0} \omega_{i,\ell}
$,
which is exactly the $\ell_0$ norm of $\mathbf{z}_i$, and meets the sparsity constraint $L_i<L_0$. Fig.~\ref{fig:model} illustrates the SDLGM. 

\subsection{The Generative Process}
\label{sec:generative}
Now we define the details of the SDLGM model with the generative process.  Formally we draw an observation $\mathbf{x}_i$ in the following manner.
\begin{itemize}
\item Draw $L_0$ number of i.i.d. random variables $\omega_{i,\ell} \in \{0,1\}$ from a Bernoulli distribution with a hyperparameter $0<\lambda_0<1$:
\begin{align}
\label{eq:generate_omega}
\omega_{i,\ell} \sim \text{Bernoulli}(\lambda_0)
\end{align}
Here the sparsity variable $L_i$ follows binomial distribution with hyperparameters $L_0$ and $\lambda_0$, i.e., 
\begin{align}
L_i = \sum\nolimits_{\ell=1}^{L_0} \omega_{i,\ell}; \;\;  
L_i \sim \text{Binomial}(L_0, \lambda_0). \nonumber
\end{align}
Here we do not directly sample $L_i$ from the binomial distribution. The suggested sampling process is computationally efficient due to matrix manipulation of DNNs in the inference.
\item The latent representation $\mathbf{z}_i$ is a weighted sum of $L_0$ i.i.d. random features. Each feature $\mathbf{z}_{i,\ell}$ is a K-dimensional one-hot categorical variable. They are independent, but share a common categorical distribution with the parameter $\alpha$ that lies in a $(K-1)$-dimensional simplex $\Delta^{K-1}$, i.e.
$
\sum_{k=1}^K \alpha_{k}=1; \;\; \alpha_{k}>0.
$
Here we assume a prior that gives equal probability to each dimension $\alpha_{k} = 1/K$.  The representation $\mathbf{z}_i$ is sampled as:
\begin{align}
\label{eq:generate_z}
\mathbf{z}_i = \sum\nolimits_{\ell=1}^{L_0} \omega_{i,\ell} \mathbf{z}_{i,\ell}; \;\;
\mathbf{z}_{i,\ell} \sim \text{Categorical}(\alpha)
\end{align}
\item Finally we sample the binary observation $\mathbf{x}_i \in \{0,1\}^D$ conditioned on the latent representation $\mathbf{z}_i$. Each observation has distinct Bernoulli parameters $\beta_i \in \mathbb{R}^D$, which are computed as:
\begin{align}
\beta_i = g(\mathbf{z}_i;\theta),\nonumber
\end{align}
where the function $g(\cdot)$ is represented using a deep neural network, as known as \textit{decoder}, with $\mathbf{z}_i$ as input and $\theta$ as hyperparameters. Given the computed Bernoulli parameter $\beta_{i,d}$, we can then independently draw each dimension of the observation:
\begin{align}
\label{eq:generate_x}
x_{i,d} \sim \text{Bernoulli}(\beta_{i,d}).
\end{align}
\end{itemize}
The decoder network $g$ can be flexible, but the output layer will use sigmoid activation to meet the constraint of the Bernoulli parameter, i.e., $0<\beta_{i,d}<1$. 

The generative model SDLGM presents an elegant way to clearly capture the sparsity of the latent code. In Bayesian modeling, the typical solution for sparseness is to set a prior distribution on the parameters, e.g., Gaussian prior as L2 regularization and Laplace prior as L1 regularization. Instead of these implicit ways, we introduce auxiliary variables $\omega_{i,\ell}$ to model the sparsity of the latent representation clearly. It explicitly switches on a finite number of latent features to generate an observation. In addition, the discrete latent codes are generally modeled with Bernoulli or one-hot categorical distributions in Bayesian framework. In this paper, we model them with a novel way (two sampling steps) to clearly learn the sparsity of the latent codes, shown as Equation~\ref{eq:generate_omega} and \ref{eq:generate_z} for generative process and Equation~\ref{equ:2dgumbelsoftmax} and \ref{equ:kdgumbelsoftmax} for approximate inference. With the proposed model, we can learn the sparse representation under a certain threshold on degree of sparsity. It would be especially useful in the low-resource applications, e.g. edge domain of 5G network with limited communication bandwidth, and a surveillance system with storage constraint.
In summary, the joint distribution of a single observation is computed as:
\begin{align}
&p(\mathbf{x}_i, \{\omega_{i,\ell},\mathbf{z}_{i,\ell}\}_{\ell=1}^{L_0} |\theta,\lambda_0,L_0,\alpha) \nonumber\\
&= p(\mathbf{x}_i|\{\mathbf{z}_{i,\ell},\omega_{i,\ell}\}_{\ell=1}^{L_0},\theta)
\prod\nolimits_{\ell=1}^{L_0} p(\omega_{i,\ell}|\lambda_0) p(\mathbf{z}_{i,\ell}|\alpha)
\end{align}

\subsection{Inference and Learning}
\label{sec:inference}

In this section we introduce computational challenges in fitting the sparse model SDLGM. The key inferential problem is to compute the posterior distribution of the sparse latent representation $\mathbf{z}_i=\sum\nolimits_{\ell=1}^{L_0} \omega_{i,\ell} \mathbf{z}_{i,\ell}$ given an observation $\mathbf{x}_i$:
\begin{align}
\label{equ:posterior}
&p(\{\omega_{i,\ell},\mathbf{z}_{i,\ell}\}_{\ell=1}^{L_0} | \mathbf{x_i},\theta,\alpha,\lambda_0,L_0)
\nonumber \\
&\propto
p(\{\omega_{i,\ell},\mathbf{z}_{i,\ell}\}_{\ell=1}^{L_0}|\alpha,\lambda_0)
p(\mathbf{x_i}| \{\omega_{i,\ell},\mathbf{z}_{i,\ell}\}_{\ell=1}^{L_0}, \theta)
\end{align}
The posterior distribution \eqref{equ:posterior} is intractable for exact inference due to the normalization term $p(\mathbf{x_i}|\theta,\alpha,\lambda_0,L_0)$. We approximate the intractable posterior based on the amortized variational inference (VI). 

Amortized VI \cite{Kingma2014,Rezende2014,Gershman2014,Zhang2019,Dayan1995} is a recent advance in Bayesian inference and deep learning. Instead of directly optimizing a set of variational parameters, it models the variational parameters as functions of observations, and then optimizes the parameters of the functions to approximate the posterior. In deep learning, the functions are often represented with neural networks, that map from observation space to variational parameter space. 
The inference of the SDLGM is implemented under the amortized VI framework. Note that the sparsity variables and the feature variables are discrete, thus we use Gumbel softmax distributions to approximate their non-differentiable discrete counterparts.

Based on Jensen's inequality and fully-factorized variational distributions \cite{Ghahramani2001}, the lower bound 
$\mathcal{L}$ (a.k.a.~evidence lower bound ELBO) of the log likelihood 
$\log p(\mathbf{x_i}|\theta,\alpha,\lambda_0,L_0)$ is: 
\begin{align}
\label{equ:loglikelihood}
&\mathcal{L}
= \sum\nolimits_{\ell=1}^{L_0} \mathbb{E}_{q(\mathbf{\omega}_{i,\ell})}  \left[ \log p(\mathbf{\omega}_{i,\ell}|\lambda_0)\right] \nonumber\\
&\;\;+ \sum\nolimits_{\ell=1}^{L_0} \mathbb{E}_{q(\mathbf{z}_{i,\ell})} \left[ \log p(\mathbf{z}_{i,\ell}|\alpha)\right] \nonumber\\
&\;\;- \sum\nolimits_{\ell=1}^{L_0} \mathbb{E}_{q(\omega_{i,\ell})} \left[ \log q(\omega_{i,\ell}) \right]  \nonumber\\
&\;\;- \sum\nolimits_{\ell=1}^{L_0} \mathbb{E}_{q(\mathbf{z}_{i,\ell})} \left[ \log q(\mathbf{z}_{i,\ell}) \right]
\nonumber\\ 
&\;\;+ \mathbb{E}_{q(\{\omega_{i,\ell},\mathbf{z}_{i,\ell}\}_{\ell=1}^{L_0})} 
\left[ \log p(\mathbf{x_i}|\theta, \{\omega_{i,\ell}, \mathbf{z}_{i,\ell}\}_{\ell=1}^{L_0}) \right]
\end{align}
On the right side of Equation~\eqref{equ:loglikelihood}, the combination of the first four terms is the negative KL divergence between variational and prior distributions. The last term is the variational expectation of the log likelihood of the data. Now the inference problem is converted to find the tightest possible lower bound $\mathcal{L}$, i.e. maximizing $\mathcal{L}$ w.r.t. the variational distributions.

\begin{figure*}
\centering
\includegraphics[width=14cm]{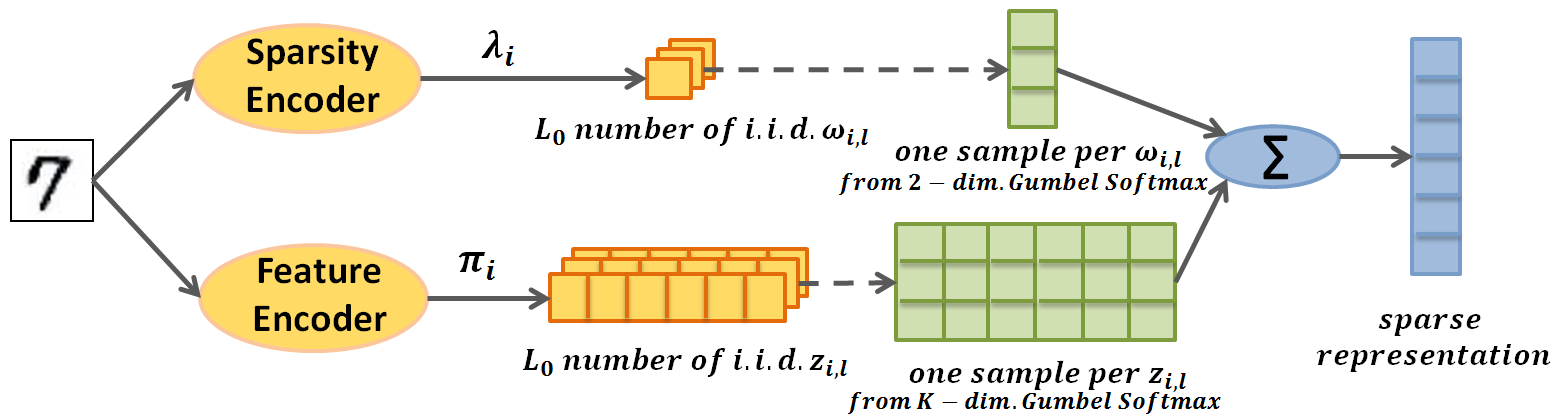}
\caption{The encoder computational graph for the amortized variational inference of the SDLGM model.}
\label{fig:encoder}
\end{figure*}
Based on the amortized VI, we use deep neural networks (DNNs) to define the variational distributions. For the sparsity variable $\omega_{i\ell}$, it is natural to assume:
\begin{align}
q(\omega_{i,\ell}) = Bernoulli(\lambda_i); \;\;
\quad \lambda_i = h(\mathbf{x}_i;\gamma),
\end{align}
where we assume that the Bernoulli parameter $0<\lambda_i<1$, one for each observation, is computed with a function $h(\mathbf{x}_i;\gamma)$, which is a DNN with $\mathbf{x}_i$ as input and $\gamma$ as hyperparameters. However the discrete variables make the corresponding layers no longer differentiable, such that the efficient computation of parameter gradients become difficult for backpropagation. A natural question to arise is: how do we find a continuous approximation to the discrete one to solve the issue. We use 2-dimensional Gumbel-softmax, a differentiable approximation to Bernoulli distribution. Gumbel-softmax distribution is defined over continuous random variables, and is introduced by \cite{Jang2017,Maddison2017,Tucker2017,Gumbel1954} to approximate categorical and Bernoulli distributions. A sample of $\omega_{i,\ell}$ from a 2-dimensional Gumbel-Softmax with parameters $\lambda_i$ and a hyperparameter $\tau \in \mathbb{R}$ (known as temperature) is defined as follows:
\begin{align}
\label{equ:2dgumbelsoftmax}
\omega_{i,\ell} &= \frac{\exp((\log(\lambda_{i,1})+\xi_{i,\ell,1})/\tau)}
{\sum_{k'=1}^2 \exp((\log(\lambda_{i,k'})+\xi_{i,\ell,k'})/\tau)}
\nonumber\\
\xi_{i,\ell,k'} &\sim \text{Gumbel}(0,1);\;\;
\lambda_{i,1} = \lambda_{i};\;\; \lambda_{i,2} = 1 - \lambda_{i}.
\end{align}

Similarly the variational distribution $q(\mathbf{z}_{i,\ell})$ of a single latent feature $\mathbf{z}_{i,\ell}$ is defined as a K-dimensional continuous Gumbel-softmax with parameters $\pi_i$ and a hyperparameter $\tau_z$:
\begin{align}
q(\mathbf{z}_{i,\ell}) = \text{Gumbel-Softmax}(\pi_i,\tau_z);\;
\pi_i = f(\mathbf{x}_i;\phi).
\end{align}
Here the function $f(\cdot)$ denotes another DNN (known as \textit{encoder}) with the observation $\mathbf{x}_i$ as input and $\phi$ as hyperparameters. Figure~\ref{fig:encoder} shows the computational graph of the encoder.
Draw a sample of $\mathbf{z}_{i,\ell}$ from the K-dimensional Gumbel-Softmax as follows:
\begin{align}
\label{equ:kdgumbelsoftmax}
z_{i,\ell,k} &= \frac{\exp((\log(\pi_{i,k})+\xi_{i,\ell,k})/\tau)}{\sum_{k'=1}^{K} \exp((\log(\pi_{i,k'})+\xi_{i,\ell,k'})/\tau)}
\nonumber\\
\xi_{i,\ell,k'} &\sim \text{Gumbel}(0,1);\;\; k = 1\ldots K
\end{align}

Given the variational distributions, we can now optimize the lower bound $\mathcal{L}$. Since we use DNNs to define the variational distributions, the variational parameters $\lambda_i$ and $\pi_i$ are deterministically computed with the global functions $h(\mathbf{x}_i;\gamma)$ and $f(\mathbf{x}_i;\phi)$ (given that $\phi$ and $\gamma$ are unknown but not random). Thus we actually do not need to optimize them with an expectation step. Instead we directly maximize $\mathcal{L}$ w.r.t. the DNNs' parameters $\theta$, $\phi$, and $\gamma$.

The expectations in the lower bound \eqref{equ:loglikelihood} are now computed with the MC estimators. The variational expectation of the log likelihood of the data is derived as:
\begin{align}
\label{equ:mc_ELL}
&\mathbb{E}_{q(\{\omega_{i,\ell},\mathbf{z}_{i,\ell}\}_{\ell=1}^{L_0})} 
\left[ \log p(\mathbf{x_i}|\theta, \{\omega_{i,\ell}, \mathbf{z}_{i,\ell}\}_{\ell=1}^{L_0}) \right]
\\
&= \frac{1}{S} \sum_{s=1}^{S} \sum_{d=1}^{D}
\left \{x_{i,d} \log \beta_{i,d}^{(s)} 
+ (1-x_{i,d}) \log \left(1-\beta_{i,d}^{(s)} \right)
\right \}, \nonumber
\end{align}
\begin{align}
\label{equ:mc_ELL_sample}
&\text{where } \beta_i^{(s)} = g(\mathbf{z}_i^{(s)}; \theta);
\;\;
\mathbf{z}^{(s)}_{i} = \sum\nolimits_{\ell=1}^{L_0} \omega_{i,\ell}^{(s)} \mathbf{z}^{(s)}_{i,\ell}; 
\nonumber\\
&\quad \quad \;\; \mathbf{z}^{(s)}_{i,\ell} \sim \text{K-d Gumbel-softmax}(\pi_i,\tau_z) \text{ with \eqref{equ:kdgumbelsoftmax}}; 
\nonumber\\
&\quad \quad \;\;
\omega_{i,\ell}^{(s)} \sim \text{2-d Gumbel-softmax}(\lambda_i,\tau) \text{ with \eqref{equ:2dgumbelsoftmax}}.
\end{align}
There are some advantages introduced by the MC estimator. For example, $\mathbf{z}_{i}$ is a weighted sum of $L_0$ continuous categorical variables in theory, then its distribution is an $L_0$-fold convolution. Since $\mathbf{z}_{i,\ell}$ is now continuous and follows Gumbel-softmax distribution, the convolution is analytically difficult to obtain. With MC estimator, we do not need to explicitly compute the probabilistic density, instead, get an approximation with a sampler. 

Next we compute Kullback-Leibler divergence KLD of variational and prior distributions of the sparsity variable. Since $\omega_{i,\ell}$ are i.i.d., the KLD is derived as:
\begin{align}
\label{equ:mc_KLL}
L_0 \left[ \lambda_i \log \left(\frac{\lambda_i}{\lambda_0} \right) + (1-\lambda_i) \log \left( \frac{1-\lambda_i}{1-\lambda_0} \right) \right].
\end{align}

Similarly we compute  $\sum\nolimits_{\ell=1}^{L_0}\text{KL}(q(\mathbf{z}_{i,\ell})||p(\mathbf{z}_{i,\ell}))$ as:
\begin{align}
\label{equ:mc_KLZ}
L_0\sum\nolimits_{k=1}^K \pi_{i,k} \log(\pi_{i,k}/\alpha_{k}).
\end{align}

We have described the computation of each term in the lower bound \eqref{equ:loglikelihood}. Now we optimize the lower bound $\mathcal{L}$ (for a single observation) w.r.t. hyperparameters based on stochastic method \cite{Hoffman2013,Kingma2014}. 
Let start with the hyperparameters $\theta$. The gradient $\nabla_{\theta} \mathcal{L}$ is approximated with the MC estimator.
\begin{align}
\label{equ:mc_gtheta}
 & \mathbb{E}_{q(\{\omega_{i,\ell},\mathbf{z}_{i,\ell}\}_{\ell=1}^{L_0})} 
\left[
\nabla_{\theta} \log p(\mathbf{x_i}|\theta, \{\omega_{i,\ell},\mathbf{z}_{i,\ell}\}_{\ell=1}^{L_0})
\right] \nonumber\\
&= \frac{1}{S} \sum_{s=1}^{S} 
\left [ \nabla_{\theta} \mathbf{x}_{i} \log g(\mathbf{z}_i^{(s)}; \theta) + \nabla_{\theta} (\mathbf{1}-\mathbf{x}_{i}) \log \left( \mathbf{1}-g(\mathbf{z}_i^{(s)}; \theta) \right)
\right ],
\end{align}
where the sampling process is the same as \eqref{equ:mc_ELL_sample}. The gradient in \eqref{equ:mc_gtheta} can be easily computed with backpropagation. Note that we actually sample $\xi$ from a fixed and known distribution $\text{Gumbel}(0,1)$, then deterministically compute the continuous Bernoulli and categorical samples. It takes effect of reparameterization, analogous to the location-scale transformation for Gaussian latent representations used in \cite{Kingma2014}, which makes the MC variance reduced. Thus $S$ can be similarly set to 1 when the mini-batch size is large enough, e.g.~128 \cite{Kingma2014}.

Next we optimize $\gamma$ and $\phi$. Their gradients are:
\begin{align}
\label{equ:mc_ggamma}
\nabla_{\gamma} \mathcal{L} &= \nabla_{\gamma}  \mathbb{E}_{q(\{\omega_{i,\ell},\mathbf{z}_{i,\ell}\}_{\ell=1}^{L_0})} 
\left[ \log p(\mathbf{x}_i|\theta, \{\omega_{i,\ell},\mathbf{z}_{i,\ell}\}_{\ell=1}^{L_0}) \right] 
\nonumber\\
& - \sum\nolimits_{\ell=1}^{L_0} \nabla_{\gamma} \text{KL}(q(\omega_{i,\ell})||p(\omega_{i,\ell}))
\end{align}
\begin{align}
\label{equ:mc_gphi}
\nabla_{\phi} \mathcal{L} &= \nabla_{\phi}  \mathbb{E}_{q(\{\omega_{i,\ell},\mathbf{z}_{i,\ell}\}_{\ell=1}^{L_0})} 
\left[ \log p(\mathbf{x}_i|\theta, \{\omega_{i,\ell},\mathbf{z}_{i,\ell}\}_{\ell=1}^{L_0}) \right] 
\nonumber\\
& - \sum\nolimits_{\ell=1}^{L_0} \nabla_{\phi} \text{KL}(q(\mathbf{z}_{i,\ell}|\pi_i,\tau_z)||p(\mathbf{z}_{i,\ell}|\alpha)) 
\end{align}
The first terms can be calculated with backpropagation and MC estimator in a similar way as \eqref{equ:mc_gtheta}. The computation of the second terms are straightforward with chain rule, e.g.
\begin{align}
\nabla_{\gamma} \text{KL} (q||p) 
= \left[ 
\log \left( \frac{\lambda_i}{\lambda_0} \right) + 
\log \left( \frac{1-\lambda_i}{1-\lambda_0} \right) + 2 \right] 
\nabla_{\gamma} \lambda_i
\end{align}
where $\nabla_{\gamma} \lambda_i$ is obtained with backpropagation. The scalar hyperparameters $\tau$ and $\tau_z$, ranging from 0 to 10 in general, can be optimized with cross-validation, or be learned with gradients that are computed similarly as $\theta$, $\gamma$ and $\phi$. Put everything together, the inference method is summarized in Algorithm~\ref{alg:sgd}. Note that the gradients described here are to maximize $\mathcal{L}$, so the SGD needs negative gradients to minimize the negative lower bound $-\mathcal{L}$.
{\SetAlgoNoLine
\begin{algorithm}
\caption{Amortized variational inference for SDLGM}
\label{alg:sgd}
\SetKwInOut{Input}{Input}
\SetKwInOut{Output}{Output}
\Input{dataset $\mathcal{D}$, latent dimension $K$, $\alpha = \{\frac{1}{K}, \ldots,\frac{1}{K}\}$, $L_0 \ll K$, $\lambda_0 = 0.5$, batch size $M=100$, sample number $S=1$, initialization of $\tau=1$, $\tau_z=1$, $\theta$, $\gamma$ and $\phi$.}
\For{$t=1$ \KwTo $T$}{
Select a minibatch $\mathcal{B}$ from the dataset $\mathcal{D}$;\\
Sample and compute $\mathbf{z}_i$ with \eqref{equ:mc_ELL_sample} for $\mathbf{x}_i \in \mathcal{B}$;\\
Compute $\mathcal{L}$ of the minibatch with \eqref{equ:mc_ELL}, \eqref{equ:mc_KLL}, \eqref{equ:mc_KLZ};\\
Compute the gradients $\nabla_{\theta,\gamma,\phi}$ with \eqref{equ:mc_gtheta}, \eqref{equ:mc_ggamma}, \eqref{equ:mc_gphi};\\
Update $\tau$, $\tau_z$, $\theta$, $\gamma$ and $\phi$ using the gradients;
}
\Output{the optimized $\tau$, $\tau_z$, $\theta$, $\gamma$ and $\phi$}
\end{algorithm}
}

\subsection{The SDLGM for Supervised Learning}

In the supervised learning setting, the SDLGM can be tailored to solve the multi-class multi-label classification issue. Assume there are $K$ possible classes in total. For an observation $\mathbf{x}_i$, let denote its labels as a binary vector $\mathbf{y}_i \in \{0,1\}^K$. An observation can have a varying number of labels, i.e., $\sum\nolimits_{k=1}^K \mathbf{y}_{i,k} = L_i$ is not fixed. Here the SDLGM has three variants, which employ different modeling strategies. 
 
\noindent \textbf{Discriminative modeling}. We directly estimate the target probability  $p(\mathbf{y}_i|\mathbf{x}_i,\alpha,\lambda_0,L_0,\tau,\tau_z,\gamma,\phi)$ without modeling the generative process of the data.
Technically, it is the encoder part described in the above section, and the variable $\mathbf{z}_i$ is now the observable labels, i.e.~$z_i\equiv y_i$.  Since $\mathbf{y}_i$ is observed, we can derive the parameters $\gamma$ and $\phi$ of the encoder via optimizing the log likelihood of the data $\sum\nolimits_{i=1}^N \log p(\mathbf{y}_i|\mathbf{x}_i,\alpha,\lambda_0,L_0,\tau,\tau_z,\gamma,\phi)$. The gradients are similarly computed with the equations \eqref{equ:mc_ggamma} and \eqref{equ:mc_gphi}.

\noindent \textbf{Generative modeling}. We model the generative process of the data, which is the same as Fig.~\ref{fig:model}. Again, the variable $\mathbf{z}_i$ is now the observable labels, $z_i\equiv y_i$. The inference is thus similar as the last section. The major difference is that there is now an extra loss due to the observed labels. After learning the parameters of the model, we use the encoder network to predict the labels $y_*$ of a test data.

\noindent \textbf{Conditional generative modeling}. We assume that the observation $x_i$ is conditioned on not only its observable labels $y_i$, but also some latent features. Technically, a part of the variable $z_i$ in the SDLGM is observed (denoted as $y_i$), and the rest is still unobservable. Different from the conditional VAE \cite{Sohn2015}, the SDLGM models $q(z_i,y_i|x_i)$ with the encoder network, rather than $q(z_i|x_i,y_i)$, since the learning tasks are not only generating the observations, but also classifying the multi-class multi-labels.

In theory, the discriminative SDLGM would show better performance than the generative one since it directly optimizes the target probability \cite{Ng2002}. The generative SDLGM would be good at generality due to the additional flexibility of modeling the data generative process. Intuitively the conditional generative SDLGM might achieve more accurate classification results than the generative SDLGM, as its generative process can be more reasonable: the observations depend on not only their labels, but also some hidden features and random effects.

\section{Empirical Analysis}
\label{sec:exp}

In the experiment session, we evaluate performance of the SDLGM with multiple datasets for different learning tasks. The proposed method is first verified for an unsupervised learning task to train sparse image representation with the popular benchmark dataset MNIST. We then test the SDLGM for a supervised learning task about multi-class multi-label classification with the datasets: Birds, Emotions and CAL500. Finally the SDLGM is applied to predict appearance times of items in an image with a  digit dataset created from MNIST.

\subsection{Unsupervised Learning Task}
We start with the representation learning task. The SDLGM is compared with state-of-the-art methods, including Gumbel-Softmax, ST Gumbel-Softmax, Score-Function (SF), DARN, MuProp, Straight-Through (ST), and Slope-Annealed ST \cite{Jang2017}. The performance of the methods is measured with negative variational lower bound of the log-likelihood (lower is better), which is a commonly used criterion in the literature.  
In the experiments, we follow the similar settings as \cite{Jang2017}. The popular image dataset MNIST is used. The encoder is defined as a Multi-Layer Perceptron (MLP) network with three hidden layers [512, 384, 256]. The input are $28 \times 28$ image data. The outputs are the logits of the distributions of 200d. The sparsity part is another MLP net with 2 hidden layers [256, 64], and with 2d outputs for parameters of continuous binomial distribution. The hyperparameters $\tau$ and $\tau_z$ are initially set as $\tau_0 = 1.0$ and are learned with the stochastic backpropagation. The hyperparameter $L_0$ is set to 40.  The decoder is defined as a reverse MLP network [256, 384, 512]. The outputs are 784d vectors, which are the Bernoulli logits of the pixels.

\begin{table}
\centering
\caption{Unsupervised learning with the SDLGM. Comparison with the baselines reported in \cite{Jang2017}. The performance is measured with negative ELBO (lower is better).}
\label{tab:elbo}
\small
\begin{tabular}{l|c|c}
\hline
\multirow{2}{*}{Methods} & \multicolumn{2}{c}{Negative ELBO} \\ 
\cline{2-3}
 & Categorical  & Bernoulli \\
\hline
SF & 110.6 & 112.2 \\
DARN & 128.8 & 110.9 \\
MuProp & 107.0 & 109.7 \\
ST & 110.9 & 116.0 \\
Annealed ST & 107.8 & 111.5 \\
Gumbel-Softmax & 101.5 & 105.0 \\
ST Gumbel-Softmax & 107.8 & 111.5 \\
\hline 
\multirow{2}{*}{Our SDLGM}   & \multicolumn{2}{c}{99.05 $\pm$ 1.07 (test)}  \\ 
  & \multicolumn{2}{c}{99.58 $\pm$ 1.08 (training)}  \\
\hline
\end{tabular}
\vspace{-0.3cm}
\end{table}
We first quantitatively demonstrate the performance of the proposed method with the commonly used measurement: the negative variational lower bound (lower is better). 
We compare the SDLGM with multiple baselines, shown as Table~\ref{tab:elbo}. For a fair comparison, we directly use the results reported in \cite{Jang2017}. More details of the baselines refer to \cite{Jang2017}. The compared methods model the latent representations as 200d Bernoulli variables or 20 number of 10d categorical variables (i.e., the same number of latent logits as ours). We rerun the experiment 10 times, and report the mean. For each time, we train the model with 10000 iterations, and get the average ELBO over last 200 iterations. One can see that our sparse method achieves significantly better performance than the recent baselines. The SDLGM finds the tighter variational lower bound. 

The proposed method has extra flexibility via learning additional sparsity variables. A natural question then arises that: is the method computationally expensive? We empirically evaluate the time cost, and compare with Jang et al.’s method. We run the methods at the same computing environment and with the same hidden structures ([512,384,256] for the feature encoder, and [256,384,512] for the decoder), the same batch size (100). Experimentally, our method uses 0.013s per iteration, while Jang et al.’s method uses 0.009s. The total time costs would be comparable, since we have less iterations (10k vs 100k). The SDLGM can learn the sparse representations in an efficient way since: 
\begin{enumerate}
\item the distributions of the data sparsity are also represented with neural networks and are properly embedded in the deep generative modeling framework, thus can be efficiently learned with backpropagation. 
\item the sparsity is represented as a function of the observation, and thus can share some complicated low layers with the feature encoder. 
\item it largely reduces the number of parameters. 
\end{enumerate}

\begin{figure}[!ht]
\centering
\begin{tabular}{c c }
\includegraphics[height=4cm]{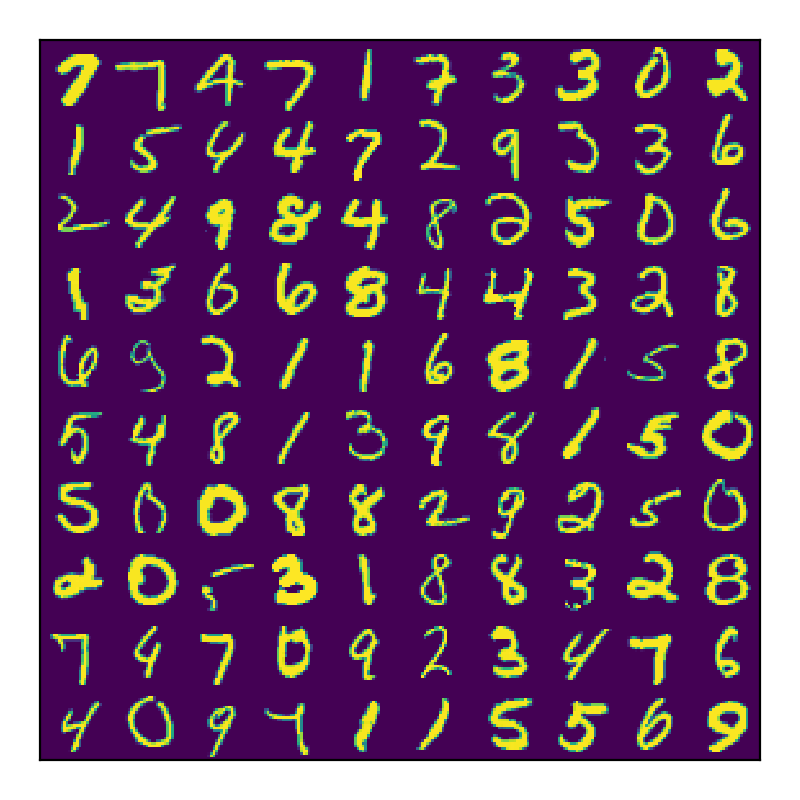}  \hspace{-0.4cm}  &
\includegraphics[height=4cm]{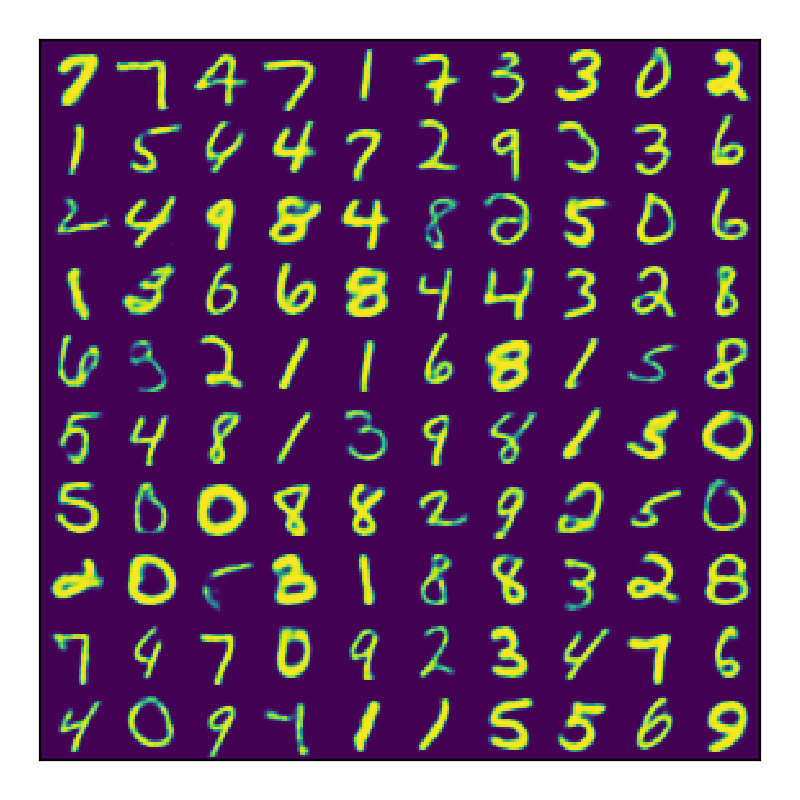}   
\end{tabular}
\vspace{-0.3cm}
\caption{The original images (left) and the SDLGM-reconstructed images (right).}
\label{fig:recon}
\vspace{-0.3cm}
\end{figure}

To qualitatively analyze the performance of the SDLGM, we use it to reconstruct images. Figure~\ref{fig:recon} shows some random examples: the original images (left), and the reconstructed ones (right). One can find that the reconstructed images are very clear, and almost the same as the original ones. 

\begin{figure*}[!ht]
\centering
\begin{tabular}{c c c }
\includegraphics[height=4.8cm]{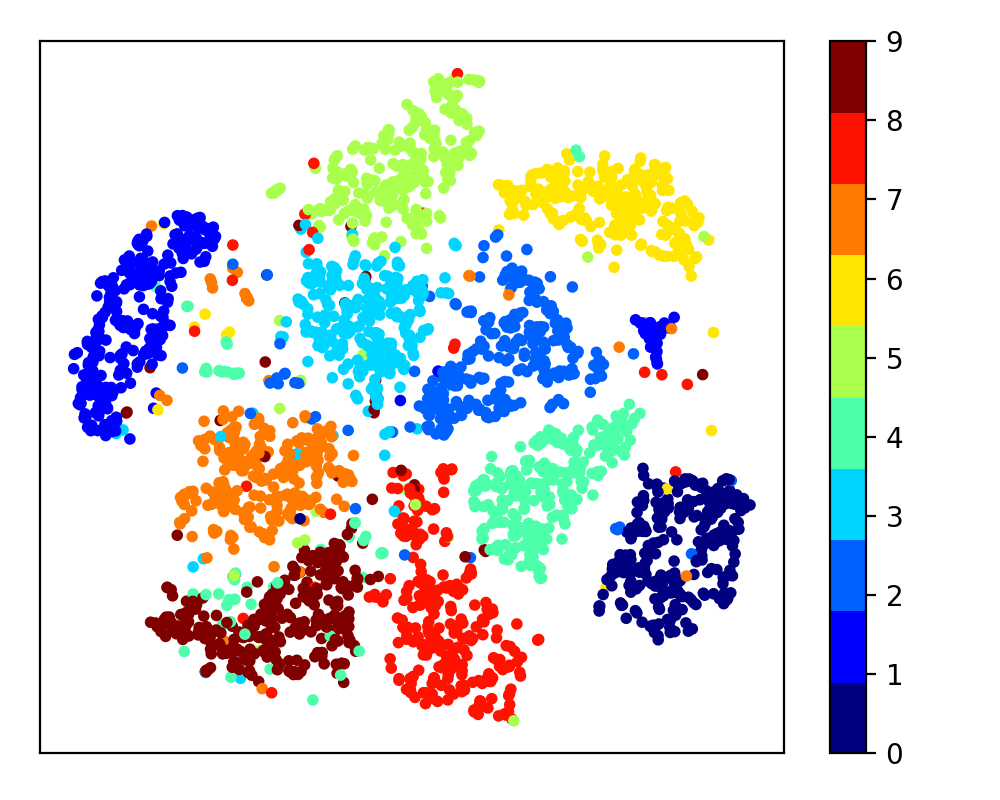}  \hspace{-0.4cm}  &
\includegraphics[height=5cm]{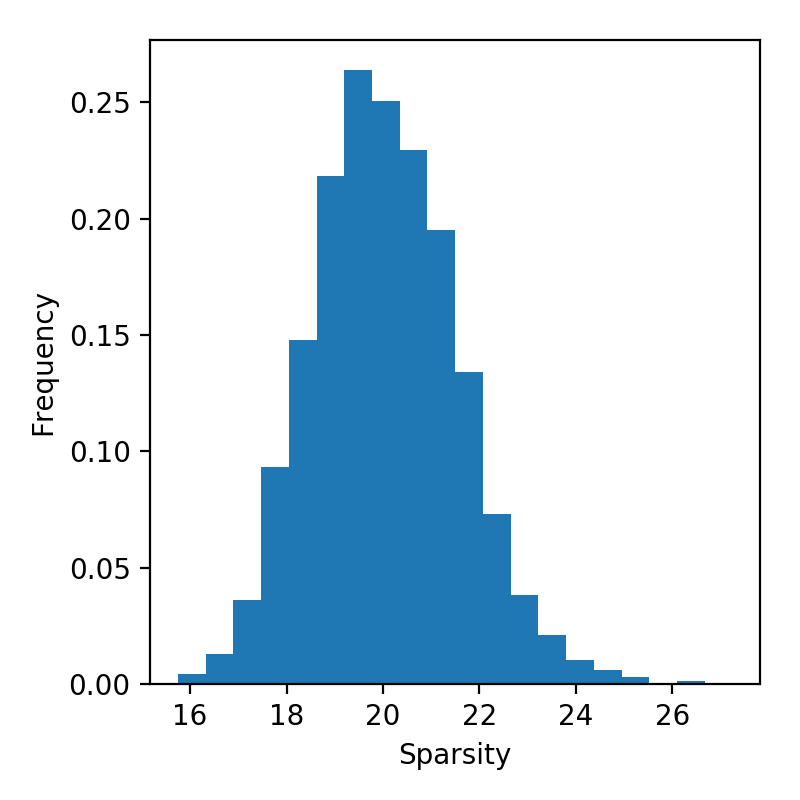}    \hspace{-0.4cm}  &
\includegraphics[height=5cm]{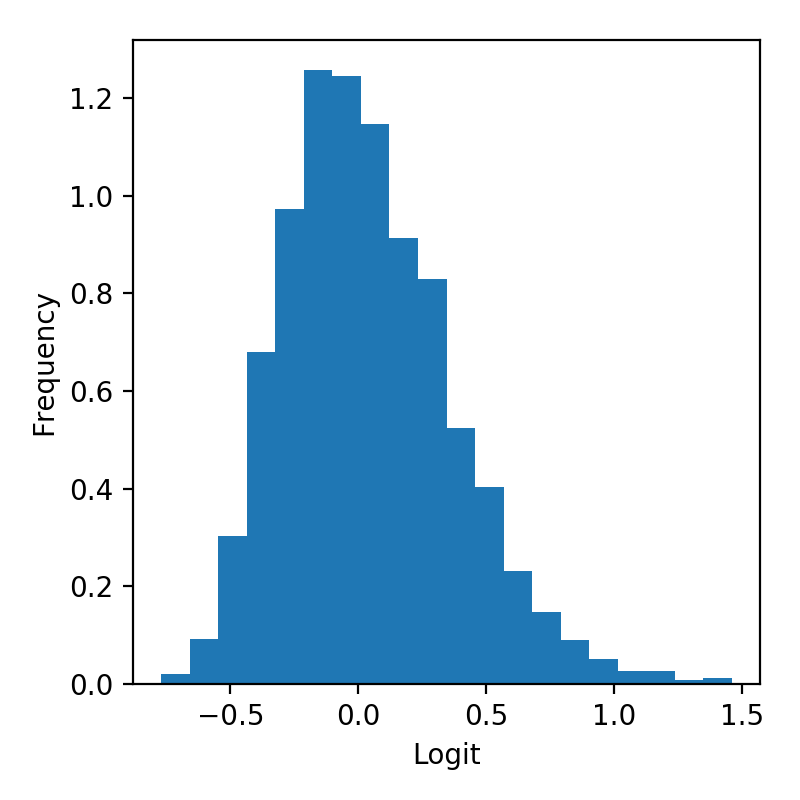}
\end{tabular}
\caption{Analysis of the estimated sparse representations of 3000 test images. Left: the learned latent structure of the digit images. One color denotes one digit (0-9). Middle: distribution of the learned sparsity $L_i$. Right: distribution of the learned sparsity logits (output of the MLP for $h(\mathbf{x}_i|\gamma)$).}
\label{fig:sparse}
\end{figure*}

\begin{figure*}[!ht]
\centering
\begin{tabular}{c c c }
\includegraphics[height=5cm]{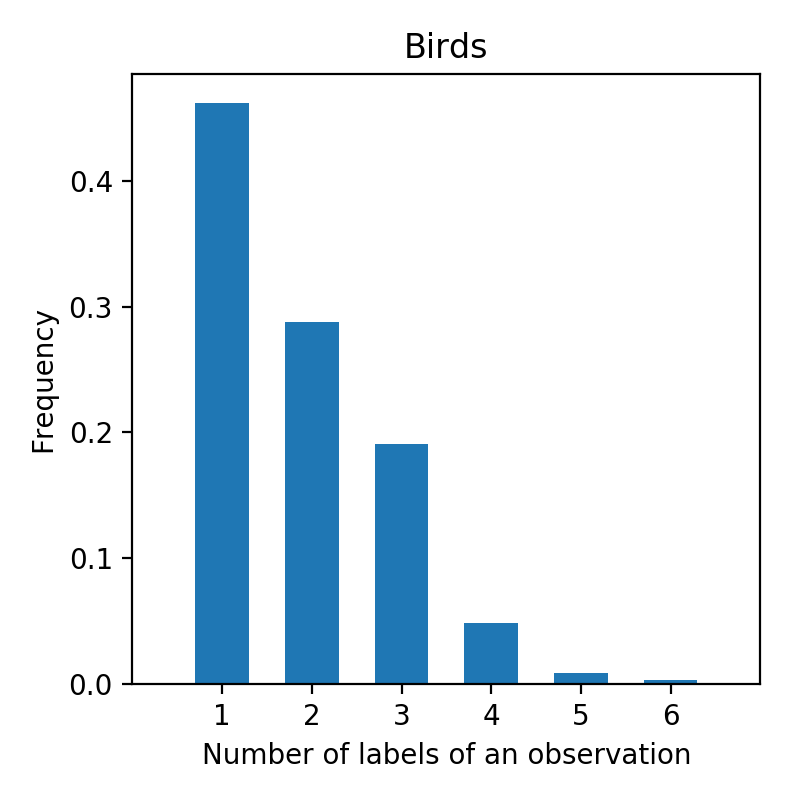}    &
\includegraphics[height=5cm]{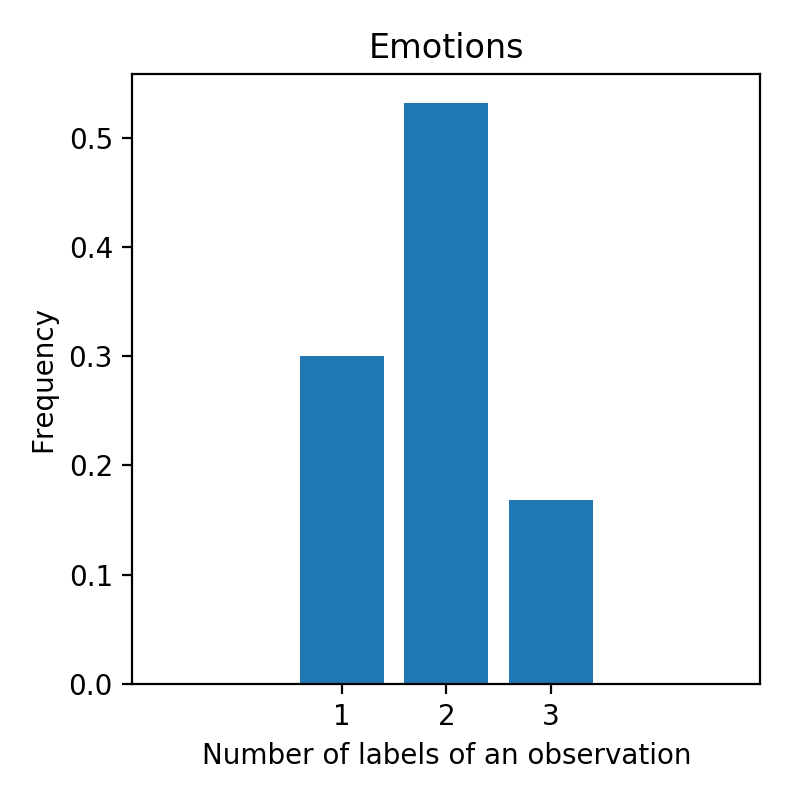}    &
\includegraphics[height=5cm]{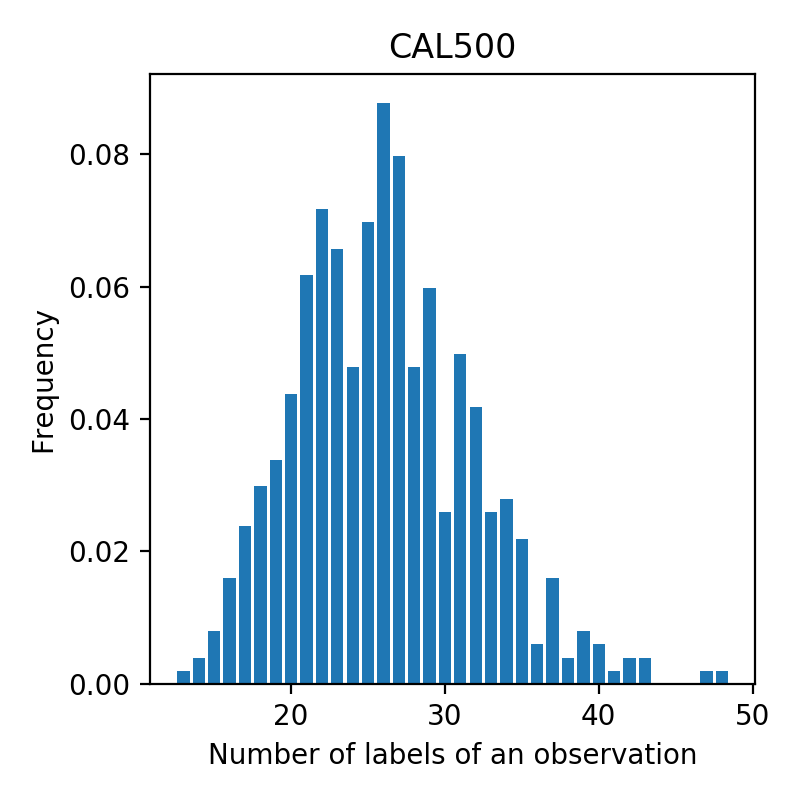}
\end{tabular}
\caption{Label sparsity of the multi-class multi-label classification task.}
\label{fig:num_labels}
\end{figure*}
Furthermore we analyze the learned sparse discrete representations. Figure~\ref{fig:sparse} shows the results with 3000 test images. The left panel illustrates the distributions of the sparse representations. 
One can find that the images of different digits (0-9) are mapped into the embedding space with different distributions. The sparse codes $\mathbf{z}$ intrinsically capture the visual concepts of the images, and clearly reveal the latent structures of the data. The two histograms represent the distributions of the sparsity $L_i$ (right panel) and of the sparsity logits (left panel, outputs of the MLP for $h(\mathbf{x}_i|\gamma)$). They are exactly close to binomial distributions.

\subsection{Multi-class Multi-Label Classification}
We also validate the performance of the proposed method in the supervised learning setting, i.e.~multi-class multi-label classification.
The experiment is performed with three datasets: Birds, Emotions and CAL500 (\url{http://mulan.sourceforge.net/datasets-mlc.html}). 
To have a fair comparison with state-of-the-art methods, we follow the experiment settings in \cite{Martins2016}. We remove the observations without any labels, and normalize the attributes to have zero mean and unit variance. 
The datasets include multiple classes: 19 (Birds), 6 (Emotions) and 174 (CAL500). Figure~\ref{fig:num_labels} shows the sparsity of the labels. We can find that the number of labels associated with an observation is not fixed, and relatively sparse considering the possible states.
For hyperparameter selection we performed a 5-fold cross validation. The evaluated hyperparameters are $L_0$, the probability thresholds $k \in \left\{ 0.05 \times n \right\}_{n=1}^{10}$ above which a label is predicted to be 1, the temperatures $\tau \in \left\{ 2.5 + 0.1 \times n\right\}_{n=0}^{10}$. \\
The selected hyperparameters are: $\tau = 3$, $L_0 = 10$ and $k = 0.4$ for Birds, $\tau = 3$, $L_0 = 2$ and $k = 0.2$ for Emotions, $\tau = 3$, $L_0 = 50$ and $k = 0.2$ for CAL500. We rerun the experiments 5 times and report the average, shown as Table~\ref{tab:results}. The three variants of the SDLGM are denoted as: SDLGM-Dis (discriminative version), SDLGM-Gen (generative version), and SDLGM-Con (conditional generative one).  Compared with the baselines, the SDLGM achieve better performance in most settings. In particular, the SDLGM with conditional generative modeling provides good predictions. The reason might be the good generality and the reasonable generative assumptions of the variant. The discriminative SDLGM achieves comparable results. The tendency coincides with the theoretical analysis in the above section.
\begin{table*}[!ht]
\centering
\caption{Micro and macro-averaged F1 scores for the logistic, softmax, sparsemax and SDLGM methods on the benchmark datasets: Birds, Emotions and CAL500.}
\label{tab:results}
\begin{tabular}{p{2cm}|>{\centering\arraybackslash}p{1.4cm}|>{\centering\arraybackslash}p{1.4cm}|>{\centering\arraybackslash}p{1.4cm}|>{\centering\arraybackslash}p{1.4cm}|>{\centering\arraybackslash}p{1.4cm}|>{\centering\arraybackslash}p{1.4cm}}
\hline
\multirow{2}{*}{Methods} & \multicolumn{2}{c}{Birds}  & \multicolumn{2}{|c}{Emotions} & \multicolumn{2}{|c}{CAL500}\\ 
\cline{2-7}
 & Micro F1  & Macro F1 & Micro F1  & Macro F1 & Micro F1  & Macro F1 \\
\hline
Logistic  & 45.78 & 33.77  & 66.75 & 68.56 & 48.88 & 24.49 \\
Softmax   & 48.67 & 37.06  & 67.34 & 67.51 & 47.46 & 23.51 \\ 
Sparsemax & 49.44 & 39.13  & 66.38 & 66.07 & 48.47 & $\mathbf{26.20}$ \\
\hline
SDLGM-Dis  & $\mathbf{50.75}$  & 42.31 & 70.99 & 70.86 & $\mathbf{49.58}$ & 20.49 \\ 
SDLGM-Gen  & 49.96 & 41.68 & 71.35 & 71.07 & 44.22 & 20.17 \\ 
SDLGM-Con  & 50.67 & $\mathbf{43.15}$ & $\mathbf{71.45}$ & $\mathbf{71.22}$ & 45.40 & 25.74\\ 
\hline
\end{tabular}
\end{table*}

\subsection{Sparsity Prediction in Image Classification}
Typically, image classification focuses on assigning a single class per image, e.g. 0-9 digits in the MNIST data. However, the real images often include multiple digits, and the numbers of appearances of the digits are varying. Although it appears to be similar to multi-label classification, it differs in which the labels can be repetitive, i.e.~one digit may appear multiple times. We solve the problem with the SDLGM to predict the appearance times (sparsity). For this end, we created a new dataset where each image is a composition of 5 columns, and each column is a random digit or an empty space. We generate 5k images for training, and 1k for test. The output of the SDLGM for each image is a 10D vector. The value of each dimension specifies how many times the corresponding digit appears in the sample. If the value is zero, then there is not the digit in the sample. We compare the proposed method with random forest in terms of mean-squared-error (MSE). On the test data, the MSE of the SDLGM is 0.05, while that of the random forest is 0.16. The proposed method achieves better performance. To illustrate the results, we randomly select some test images associated with the prediction results, shown as Table~\ref{tab:imageclass}. One can find that most of the predictions, marked as green, are correct. For example, the SDLGM predicts: the digits `3' and `4' occur one time respectively in the first image, and the digit `4' occurs two times in the second image (the top row of Table~\ref{tab:imageclass}). The experiments reveal that: (1) learning degree of sparsity is a generic task and can be used to formulate different problems; (2) the proposed method SDLGM offers an effective way to model the degree of sparsity explicitly and effectively.
\begin{table*}[!ht]
\centering
\caption{Example images illustrate predictions of the SDLGM on the number of appearances of digits. The green marks the correct predictions.}
\label{tab:imageclass}
\begin{tabular}{>{\centering\arraybackslash}m{3cm}|>{\centering\arraybackslash}m{2.5cm}||>{\centering\arraybackslash}m{3cm}|>{\centering\arraybackslash}m{2.5cm}}
\hline
Image & Prediction & Image & Prediction\\
\hline 
\includegraphics[valign=c,width=3cm]{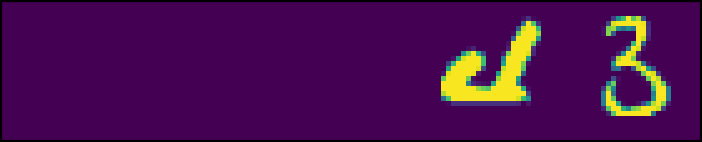} &
\parbox[m]{2.5cm}{{\color{ForestGreen} Digit `3' $ \times 1$\\Digit `4' $\times 1$}} &
\includegraphics[valign=c,width=3cm]{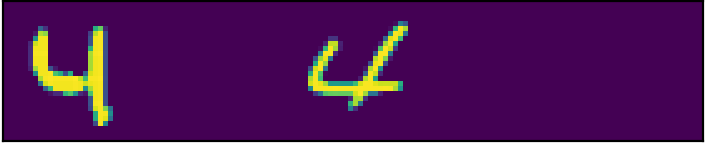} & 
\parbox[m]{2.5cm}{{\color{ForestGreen}Digit `4' $ \times 2$}}\\
\hline
\includegraphics[valign=c,width=3cm]{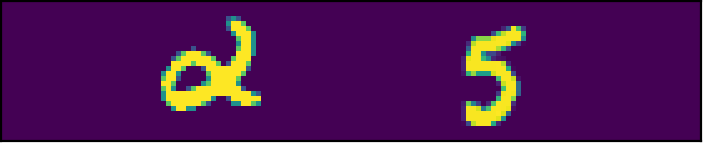} &
\parbox[m]{2.5cm}{Digit `3' $ \times 1$\\{\color{ForestGreen}Digit `5' $\times 1$}} &
\includegraphics[valign=c,width=3cm]{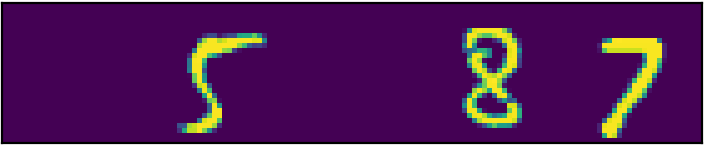} & 
\parbox[m]{2.5cm}{{\color{ForestGreen}Digit `5' $ \times 1$\\Digit `7' $\times 1$}\\Digit `9' $\times 1$}\\
\hline
\includegraphics[valign=c,width=3cm]{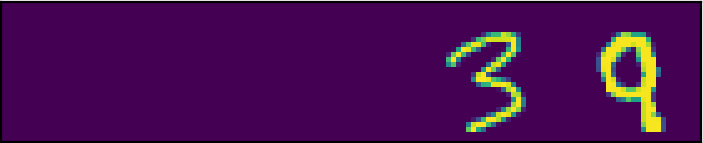} & 
\parbox[m]{2.5cm}{{\color{ForestGreen}Digit `3' $ \times 1$\\Digit `9' $\times 1$}} &
\includegraphics[valign=c,width=3cm]{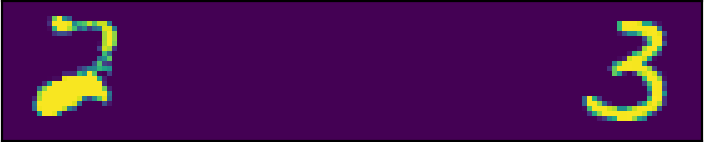} & 
\parbox[m]{2.5cm}{{\color{ForestGreen}Digit `2' $ \times 1$\\Digit `3' $\times 1$}}\\
\hline
\end{tabular}
\end{table*}

\section{Conclusion}
\label{sec:conclusion}
In this paper we present a sparse deep latent generative model, which effectively learns sparse discrete representations of the data. Instead of the commonly used regularization techniques, our model introduces an additional random variable to each observation to explicitly model the sparsity of the latent codes. With the proposed method, we are able to learn sparse representation under a certain threshold of sparsity. For inference, we develop an amortized variational method based on recent advances in Bayesian learning. The SDLGM is also tailored for the multi-class multi-label classification issue. The experiment results on multiple datasets show the benefits of the proposed model. Our work provides interesting avenues for future work, e.g. adapting the SDLGM to learn sparse attention for text and visual data, investigating the effectiveness of the proposed method for large scale datasets. 
In addition, we will extend the SDLGM to learn sparse representations of  measurement data of telecommunication networks (collected in the H2020 MonB5G project). The measurement data is important for the network operators to monitor the status of the networks, however it introduces significant storage and communication overhead to the network management platforms. Compressing the data to sparse representations improves data storage efficiency, computation cost and communication constraints. The proposed method SDLGM provides an effective solution for data compression with extra flexibility. In particular, it allows clear definition of sparsity degree: the number of non-zero features of a representation cannot be larger than a pre-defined threshold $L_0$. We will validate the effectiveness of the the SDLGM method in the network measurement data, especially with respect to the tradeoff between the performance of network status/KPIs prediction and the sparsity of the data representations.

\bibliographystyle{IEEEtran}
\bibliography{sparsity}

\begin{thebibliography}{10}
\providecommand{\url}[1]{#1}
\csname url@samestyle\endcsname
\providecommand{\newblock}{\relax}
\providecommand{\bibinfo}[2]{#2}
\providecommand{\BIBentrySTDinterwordspacing}{\spaceskip=0pt\relax}
\providecommand{\BIBentryALTinterwordstretchfactor}{4}
\providecommand{\BIBentryALTinterwordspacing}{\spaceskip=\fontdimen2\font plus
\BIBentryALTinterwordstretchfactor\fontdimen3\font minus
  \fontdimen4\font\relax}
\providecommand{\BIBforeignlanguage}[2]{{%
\expandafter\ifx\csname l@#1\endcsname\relax
\typeout{** WARNING: IEEEtran.bst: No hyphenation pattern has been}%
\typeout{** loaded for the language `#1'. Using the pattern for}%
\typeout{** the default language instead.}%
\else
\language=\csname l@#1\endcsname
\fi
#2}}
\providecommand{\BIBdecl}{\relax}
\BIBdecl

\bibitem{Kingma2014}
D.~Kingma and M.~Welling, ``Auto-encoding variational bayes,'' in \emph{ICLR},
  2014.

\bibitem{Rezende2014}
D.~J. Rezende, S.~Mohamed, and D.~Wierstra, ``Stochastic backpropagation and
  approximate inference in deep generative models,'' in \emph{ICML}, 2014.

\bibitem{Gregor2014}
K.~Gregor, I.~Danihelka, A.~Mnih, C.~Blundell, and D.~Wierstra, ``Deep
  autoregressive networks,'' in \emph{ICML}, 2014.

\bibitem{Sohn2015}
K.~Sohn, H.~Lee, and X.~Yan, ``Learning structured output representation using
  deep conditional generative models,'' in \emph{NIPS}, 2015.

\bibitem{Jang2017}
E.~Jang, S.~Gu, and B.~Poole, ``Categorical reparameterization with
  gumbel-softmax,'' in \emph{ICLR}, 2017.

\bibitem{Miao2016}
Y.~Miao, L.~Yu, and P.~Blunsom, ``Neural variational inference for text
  processing,'' in \emph{ICML}, 2016.

\bibitem{Walker2016}
J.~Walker, C.~Doersch, A.~Gupta, and M.~Hebert, ``An uncertain future:
  Forecasting from static images using variational autoencoders,'' in
  \emph{ECCV}, 2016.

\bibitem{Xu2018}
H.~Xu, W.~Chen, N.~Zhao, Z.~Li, J.~Bu, Z.~Li, Y.~Liu, Y.~Zhao, D.~Pei, Y.~Feng,
  J.~Chen, Z.~Wang, and H.~Qiao, ``Unsupervised anomaly detection via
  variational auto-encoder for seasonal kpis in web applications,'' in
  \emph{WWW}, 2018.

\bibitem{Yan2016}
X.~Yan, J.~Yang, K.~Sohn, and H.~Lee, ``Attribute2image: Conditional image
  generation from visual attributes,'' in \emph{ECCV}, 2016.

\bibitem{Yeh2016}
R.~Yeh, Z.~Liu, D.~Goldman, and A.~Agarwala, ``Semantic facial expression
  editing using autoencoded flow,'' 2016.

\bibitem{Maddison2017}
C.~Maddison, A.~Mnih, and Y.~W. Teh, ``The concrete distribution: A continuous
  relaxation of discrete random variables,'' in \emph{ICLR}, 2017.

\bibitem{Tucker2017}
G.~Tucker, A.~Mnih, C.~Maddison, D.~Lawson, and J.~Sohl-Dickstein, ``Rebar:
  Low-variance, unbiased gradient estimates for discrete latent variable
  models,'' in \emph{NIPS}, 2017.

\bibitem{Denker1991}
J.~Denker and Y.~LeCun, ``Transforming neural-net output levels to probability
  distributions,'' in \emph{NIPS}, 1991.

\bibitem{Buntine1991}
W.~Buntine and A.~Weigend, ``Bayesian back-propagation,'' \emph{Complex
  Systems}, vol.~5, 1991.

\bibitem{Neal1993}
R.~Neal, ``Bayesian learning via stochastic dynamics,'' in \emph{NIPS}, 1993.

\bibitem{Hoffman2013}
M.~Hoffman, D.~Blei, C.~Wang, and J.~Paisley, ``Stochastic variational
  inference,'' \emph{JMLR}, pp. 1303--1347, 2013.

\bibitem{Welling2011}
M.~Welling and Y.~W. Teh, ``Bayesian learning via stochastic gradient langevin
  dynamics,'' in \emph{ICML}, 2011.

\bibitem{Graves2011}
A.~Graves, ``Practical variational inference for neural networks,'' in
  \emph{NIPS}, 2011.

\bibitem{Blundell2015}
C.~Blundell, J.~Cornebise, K.~Kavukcuoglu, and D.~Wierstra, ``Weight
  uncertainty in neural networks,'' in \emph{ICML}, 2015.

\bibitem{Kingma2014b}
D.~Kingma, D.~Rezende, S.~Mohamed, and M.~Welling, ``Semi-supervised learning
  with deep generative models,'' in \emph{NIPS}, 2014.

\bibitem{Kipf2016}
T.~Kipf and M.~Welling, ``Variational graph autoencoders,'' in
  \emph{Proceedings of the Bayesian Deep Learning Workshop}, 2016.

\bibitem{Pu2016}
Y.~Pu, Z.~Gan, R.~Henao, X.~Yuan, C.~Li, A.~Stevens, and L.~Carin,
  ``Variational autoencoder for deep learning of images, labels and captions,''
  in \emph{NIPS}, 2016.

\bibitem{Lee2008}
H.~Lee, C.~Ekanadham, and A.~Ng, ``Sparse deep belief net model for visual area
  v2,'' in \emph{NIPS}, 2008.

\bibitem{Ranzato2008}
M.~A. Ranzato, Y.-L. Boureau, and Y.~LeCun, ``Sparse feature learning for deep
  belief networks,'' in \emph{NIPS}, 2008.

\bibitem{Kavukcuoglu2010}
K.~Kavukcuoglu, M.~A. Ranzato, and Y.~LeCun, ``Fast inference in sparse coding
  algorithms with applications to object recognition,'' 2010.

\bibitem{Makhzani2014}
A.~Makhzani and B.~Frey, ``K-sparse autoencoders,'' in \emph{ICLR}, 2014.

\bibitem{Arpit2016}
D.~Arpit, Y.~Zhou, H.~Ngo, and V.~Govindaraju, ``Why regularized auto-encoders
  learn sparse representation,'' in \emph{ICML}, 2016.

\bibitem{Glorot2011}
X.~Glorot, A.~Bordes, and Y.~Bengio, ``Deep sparse rectifier neural networks,''
  in \emph{AISTATS}, 2011.

\bibitem{Goodfellow2013}
I.~Goodfellow, D.~Warde-Farley, M.~Mirza, A.~Courville, and Y.~Bengio, ``Maxout
  networks,'' in \emph{ICML}, 2013.

\bibitem{Martins2016}
A.~Martins and R.~Astudillo, ``From softmax to sparsemax: A sparse model of
  attention and multi-label classification,'' in \emph{ICML}, 2016.

\bibitem{Molchanov2017}
D.~Molchanov, A.~Ashukha, and D.~Vetrov, ``Variational dropout sparsifies deep
  neural networks,'' in \emph{ICML}, 2017.

\bibitem{Louizos2018}
C.~Louizos, M.~Welling, and D.~Kingma, ``Learning sparse neural networks
  through l0 regularization,'' in \emph{ICLR}, 2018.

\bibitem{Srinivas2016}
S.~Srinivas, A.~Subramanya, and V.~Babu, ``Training sparse neural networks,''
  2016.

\bibitem{Li2018}
W.~Li, J.~Mao, Y.~Zhang, and S.~Cui, ``Fast similarity search via optimal
  sparse lifting,'' in \emph{NIPS}, 2018.

\bibitem{Jeong2018}
Y.~Jeong and H.~O. Song, ``Efficient end-to-end learning for quantizable
  representations,'' in \emph{ICML}, 2018.

\bibitem{Paria2020}
B.~Paria, C.~Yeh, I.~Yen, N.~Xu, P.~Ravikumar, and B.~Poczos, ``minimizing
  flops to learn efficient sparse representations,'' in \emph{ICLR}, 2020.

\bibitem{Gershman2014}
S.~J. Gershman and N.~D. Goodman, ``Amortized inference in probabilistic
  reasoning,'' \emph{CogSci}, vol.~36, 2014.

\bibitem{Zhang2019}
C.~Zhang, J.~Butepage, H.~Kjellstrom, and S.~Mandt, ``Advances in variational
  inference,'' \emph{Pattern Analysis and Machine Intelligence}, 2019.

\bibitem{Dayan1995}
P.~Dayan, G.~Hinton, R.~Neal, and R.~Zemel, ``The helmholtz machine,''
  \emph{Neural computation}, vol.~7, no.~5, 1995.

\bibitem{Ghahramani2001}
Z.~Ghahramani and M.~Beal, ``Propagation algorithms for variational bayesian
  learning,'' in \emph{NIPS}, 2001.

\bibitem{Gumbel1954}
E.~Gumbel, \emph{Statistical theory of extreme values and some practical
  applications: a series of lectures}.\hskip 1em plus 0.5em minus 0.4em\relax
  U. S. Govt. Print. Office, 1954.

\bibitem{Ng2002}
A.~Ng and M.~Jordan, ``On discriminative vs generative classifiers a comparison
  of logistic regression and naive bayes,'' in \emph{NIPS}, 2002.

\end{thebibliography}

\end{document}